\documentclass[letterpaper, 10 pt, conference]{ieeeconf}  

\IEEEoverridecommandlockouts                              

\usepackage{graphicx}
\usepackage{amsmath,amssymb,amsfonts}
\usepackage{algpseudocode,algorithm}
\usepackage{subcaption}
\usepackage{makecell}
\usepackage{cite}
\usepackage{url}
\usepackage{multirow}

\title{\LARGE \bf
CubeDAgger: Interactive Imitation Learning for Dynamic Systems with Efficient yet Low-risk Interaction
}

\author{Taisuke Kobayashi$^{1}$
\thanks{\copyright 2026 IEEE.  Personal use of this material is permitted.  Permission from IEEE must be obtained for all other uses, in any current or future media, including reprinting/republishing this material for advertising or promotional purposes, creating new collective works, for resale or redistribution to servers or lists, or reuse of any copyrighted component of this work in other works.}
\thanks{*This research was supported by ``Strategic Research Projects'' grant from ROIS (Research Organization of Information and Systems) and JST, CRONOS, Japan Grant Number JPMJCS24K6.}
\thanks{$^{1}$T. Kobayashi is with the National Institute of Informatics (NII) and with The Graduate University for Advanced Studies (SOKENDAI),
        2-1-2 Hitotsubashi, Chiyoda-ku, Tokyo, 101-8430, Japan
        {\tt\small kobayashi@nii.ac.jp}}%
}

\begin{document}

\maketitle
\thispagestyle{empty}
\pagestyle{empty}

\begin{abstract}

Interactive imitation learning makes an agent's control policy robust by stepwise supervisions from an expert.
The recent algorithms mostly employ expert-agent switching systems to reduce the expert's burden by limitedly selecting the supervision timing.
However, this approach is useful only for static tasks; in dynamic tasks, timing discrepancies cause abrupt changes in actions, losing the robot's dynamic stability.
This paper therefore proposes a novel method, named CubeDAgger, which improves robustness with less dynamic stability violations even for dynamic tasks.
The proposed method is designed on a baseline, EnsembleDAgger, with three improvements.
The first adds a regularization to explicitly activate the threshold for deciding the supervision timing.
The second transforms the expert-agent switching system to an optimal consensus system of multiple action candidates.
Third, autoregressive colored noise is injected to the agent's actions for time-consistent exploration.
These improvements are verified by simulations, showing that the trained policies are sufficiently robust while maintaining dynamic stability during interaction.
Finally, real-robot scooping experiments with a human expert demonstrate that the proposed method can learn robust policies from scratch based on just 30 minutes of interaction.

\end{abstract}

\section{Introduction}

Interactive imitation learning (IIL) is a well-known methodology for improving policies by incrementally providing demonstration data~\cite{celemin2022interactive}.
In particular, dataset aggregation (DAgger)~\cite{ross2011reduction}, which is the focus of this study, allows an agent and an expert to share control of a target robotic system.
Conceptually, the expert takes the initiative at the beginning and gradually delegates control to the agent, who can complete the imitation without over- or under-learning.
By introducing an appropriate safety criterion for delegating control, the expert can correct the robot to the optimal trajectory before it falls into a high-risk situation~\cite{kelly2019hg,menda2019ensembledagger,cui2019uncertainty,oh2024leveraging}.
Such an error recovery motion can make the agent policy robust to accumulated approximation errors and/or disturbances.

Recent DAgger variants commonly have a method of discretely switching control authority, which causes two problems.
First, if the expert is a human, he/she may not be able to respond immediately to the switching timing, causing a delay in the teaching operation~\cite{black2024evaluation}.
Even if the switching timing is handled by him/her~\cite{kelly2019hg}, the switching timing itself would be delayed.
These delays are not particularly problematic for static tasks (that is why IIL mostly targets a static object manipulation), but they can be fatal for dynamic tasks.
For example, errouneous switching controllers cause discrete and abrupt changes in control commands to the robot, which could easily violate the robot's dynamic stability~\cite{liberzon1999basic}.

To enable IIL for dynamic tasks, this paper proposes a new Dagger variant, named CubeDAgger.
The proposed CubeDAgger (or C$^3$DAgger) has three `C' improvements on EnsembleDAgger~\cite{menda2019ensembledagger}.
First, the agent policy is explicitly \textbf{C}ontrolled so that it satisfies a specified level of safety for determining the control authority.
Second, the switching system is replaced with an optimal \textbf{C}onsensus system of multiple action candidates.
Finally, time-consistent \textbf{C}olored noise is injected into the agent's actions for efficient exploration, which also makes human decision-making less intractable.

\begin{figure}[tb]
    \centering
    \includegraphics[keepaspectratio=true,width=0.84\linewidth]{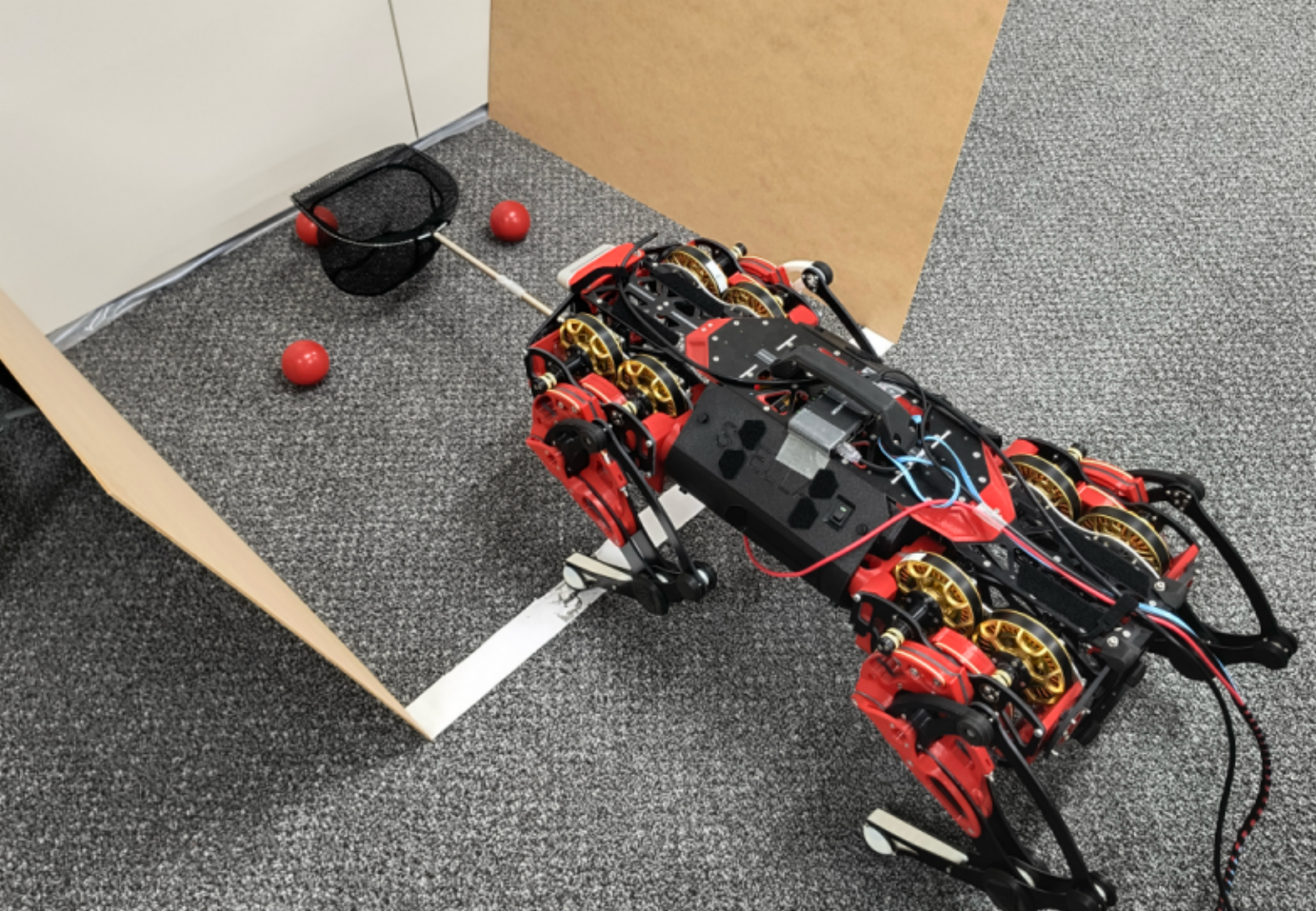}
    \caption{Scooping three balls by a net attached on a quadruped}
    \label{fig:scoop_snap}
\end{figure}

The proposed CubeDAgger is tested in three simulation environments with different dynamic characteristics.
The results show that CubeDAgger improves the robustness of the trained policies over the conventional EnsembleDAgger without sacrificing dynamic stability during data collection.
Additionally, a real-world scooping task (see Fig.~\ref{fig:scoop_snap}) is demonstrated.
Although baselines make human supervision errouneous and imitation unsatisfactory, CubeDAgger can acquire robust policies in just 30 minutes of interaction while maintaining comfortable supervision.

\subsection{Related work}

\subsubsection{DAgger family}

DAgger~\cite{ross2011reduction} (and its variants~\cite{kelly2019hg,menda2019ensembledagger,cui2019uncertainty,oh2024leveraging}) are representative of IIL.
An agent and an expert share control of a target robotic system and combine each other's actions appropriately to safely collect high-quality data.
By including the expert's actions rather than the actual actions into a dataset, the agent policy can asymptotically approach the expert one through supervised learning using it.

Recently, a switching system gained popularity as a way to combine actions of the agent and the expert.
The switching frequency that requires expert teaching is minimized, reducing the burden on the human expert~\cite{hoque2021lazydagger,hoque2022thriftydagger}.
In this case, a single human operator can teach multiple robots simultaneously, facilitating data collection on a larger scale~\cite{hoque2023fleet}.

However, such an expert-agent switching system causes a delay between the switching of control authority and the correction by the human, and when the human makes the decision to switch control authority as well~\cite{kelly2019hg}, limiting the target to systems in which the delay is ignorable~\cite{black2024evaluation}.
In addition, switching systems would violate the stability of dynamic systems because of the discrete and abrupt changes in control commands to the robot~\cite{liberzon1999basic}.
In other words, this recent approach is unsuitable for IIL on dynamic systems.

\subsubsection{DART family}

Another representative IIL methodology is disturbances for augmenting robot trajectories (DART)~\cite{laskey2017dart}.
This is a method in which the optimal (Gaussian) noise is injected into the expert's actions when demonstrating a target task, increasing the uncertainty of state transitions and adding error recovery motions into the dataset.
This noise optimization is extended to flexible representation of policies by following Bayesian inference~\cite{oh2023bayesian}.
The various trajectories diversified by the added noise are ranked to extract the better ones to improve control performance~\cite{brown2020better}.

However, as pointed out in the context of reinforcement learning (RL), stochastic exploration is inefficient~\cite{sutton2018reinforcement}.
The above DAgger is superior in terms of efficiency because it corresponds to directed exploration~\cite{wilson2021balancing}, although it is prone to exploration bias.
Therefore, it is necessary to construct a system that takes advantage of both strengths.

\subsubsection{Integration with reinforcement learning}

Although outside the scope of this study, several learning-from-demonstration methods utilizing RL have been proposed.
Inverse RL~\cite{adams2022survey} and guided RL~\cite{esser2022guided} generally specify that the goal of RL is to increase similarity with demonstration data prepared offline, and scenarios in which the expert incrementally demonstrates the target task are rare.
A method that interprets IIL as RL has emerged recently, outperforming the IIL baselines~\cite{luo2024rlif}.
However, as it involves expert interventions by means of the switching system, it faces the same problem as the recent DAgger methods.

\section{Preliminaries}

\subsection{Behavioral cloning}

At first, behavioral cloning (BC)~\cite{bain1995framework} is introduced since it is a main learning method of an agent's policy even in IIL.
Specifically, BC assumes that the target environment (i.e. a robot and surroundings) is Markovian, namely, the optimal policy, $\pi^\ast$, can be expressed only with the current state observation, $s \in \mathcal{S} \subset \mathbb{R}^{|\mathcal{S}|}$, which is sampled according to the environment's state transition probability $p_e$ (or its initial state probability $p_0$).
$\pi^\ast$ is hidden in an expert, but actions sampled from it at each state, $a \in \mathcal{A} \subset \mathbb{R}^{|\mathcal{A}|}$, can be collected.
As a result, a dataset $D = \{(s_n, a_n)\}^N_{n=1}$ with $N$ pairs is prepared in advance.

The agent's policy $\pi(a \mid s; \theta)$ with $\theta$ the trainable parameters is optimized towards $\pi^\ast$ as follows:
\begin{align}
    \theta^\ast &= \mathop{\mathrm{arg\,min}}_\theta \underbrace{\mathbb{E}_{(s_n, a_n) \sim D} \left[ -\ln \pi(a_n \mid s_n; \theta) \right]}_{\mathcal{L}(\theta)}
    \label{eq:loss_bc}
\end{align}
This problem is solved by, for example, stochastic gradient descent (e.g. AdaTerm~\cite{ilboudo2023adaterm} employed in this paper).
Note that this study assumes that $\pi$ is a diagonal normal distribution with a learnable scale for continuous action space.

\subsection{Interactive imitation learning}

In BC, $D$ is prepared and given in advance, but $D$ is intended to approximate the original expectation operation with $p_e, \pi^\ast$.
In other words, if $D$ is insufficient, the approximation error will be large and the imitation will fail eventually due to the accumulation of it.
Therefore, IIL switches the BC's problem to a new one that appends data into $D$ online, making the policy trained robust enough to the approximation error by collecting sufficient and necessary $D$.

In addition, to improve the value of expert data, how an agent would behave in the vicinity of an optimal trajectory (e.g. error recovery motions) is important.
This reduces the approximation error efficiently and provide robustness to recover from deviations from the optimal trajectory due to the accumulation of errors.
The pseudocode of DAgger family~\cite{ross2011reduction,kelly2019hg,menda2019ensembledagger,cui2019uncertainty,oh2024leveraging} with such a function is summarized in Alg.~\ref{alg:dagger}.

\begin{algorithm}[tb]
    \caption{Pseudocode of DAgger family}
    \label{alg:dagger}
    \begin{algorithmic}[1]
        \State{Initialize network parameters $\theta$, and dataset $D = \emptyset$}
        \While{not converged}
            \While{not terminated or truncated}
                \State{Get the current state $s$}
                \State{Get the expert's action $a$ over $s$}
                \State{Get the agent's action $a^\pi \sim \pi(a \mid s; \theta)$}
                \State{Determine the executed action $a^c$ from $a$ and $a^\pi$}
                \State{Transition to the next state using $a^c$}
                \State{Store experience $D = D \cup (s, a)$}
            \EndWhile
            \For{$\{ (s_i, a_i) \}_{i=1}^B \subset D$}
                \State{$\mathcal{L}(\theta) = - B^{-1} \sum_{i=1}^B \ln \pi(a_i \mid s_i; \theta)$}
                \State{Minimize $\mathcal{L}(\theta)$ by stochastic gradient descent}
            \EndFor
        \EndWhile
    \end{algorithmic}
\end{algorithm}

At Line 7 in Alg.~\ref{alg:dagger}, $a^c$, which actually acts on the environment, is determined based on $a$ demonstrated by the expert and $a^\pi$ of the agent.
There is no unique solution for determining $a^c$, so several methods have been proposed in previous studies.
In the original DAgger~\cite{ross2011reduction}, the weighted average with annealing was employed.
On the other hand, the recent trend is the following expert-agent switching system.
\begin{align}
    a^c =
    \begin{cases}
        a^\pi & (s, a^\pi) \in \mathcal{C}
        \\
        a & \mathrm{otherwise}
    \end{cases}
\end{align}
where, $\mathcal{C}$ denotes the safety set.
That is, if the agent decision $a^\pi$ at the current state $s$ is safe, $a^\pi$ is accepted; otherwise, the expert should demonstrate the correct behaviors to make the environment with robot safe enough.
The advantage of such a switching system is that the human expert does not need to demonstrate as long as the current situation is safe, and his/her burden can be greatly reduced, although this is not the case when $a$ is used to evaluate the safety.
However, this paper does not focus on this advantage, but rather on solving the problems hidden in the switching system.

\subsection{EnsembleDAgger}

As the baseline for the CubeDAgger proposed in the next section, EnsembleDAgger~\cite{menda2019ensembledagger} is introduced here.
In EnsembleDAgger, the agent evaluates the safety for switching $a$ and $a^\pi$, and the expert provides $a$ at evey time step, alleviating the delay of human decision making.

Specifically, EnsembleDAgger has $K$ ensembles of policies, $\{\pi_k(a \mid s; \theta_k)\}_{k=1}^K$, each of which is trained with eq.~\eqref{eq:loss_bc}.
Therefore, the agent has $K$ action candidates, $\{a^{\pi_k}\}_{k=1}^K$, each of which is given as the location parameter of $\pi_k$ (without stochastic exploration).
At this time, not only $a^\pi = K^{-1} \sum_k a^{\pi_k}$ but also its standard deviation $\sigma^\pi = \sqrt{K^{-1} \sum_k (a^\pi - a^{\pi_k})^2}$ can be computed.
EnsembleDAgger assumes that the current situation is safe if $a^\pi$ is close to $a$ and if $\sigma^\pi$ is enough small.
With two thresholds, $\bar{\Delta} \geq 0$ and $\bar{\sigma} \geq 0$, the safety set, $\mathcal{C}$, is defined as follows%
\footnote{In the original, both criteria are squared and the thresholds are set accordingly, but this is omitted to make it consistent with the proposal.}:
\begin{align}
    \mathcal{C} &= \{(s, a) \mid | a - a^\pi(s) | \leq \bar{\Delta} \land \sigma^\pi(s) \leq \bar{\sigma} \}
\end{align}

\section{Proposed method}

\subsection{Overview}

Since the current DAgger variants lack dynamic stability, the following three `C' modifications are proposed as \textit{CubeDAgger} to alleviate this issue.
\begin{enumerate}
    \item \textit{Controlled}:
    To reduce the errors of the hand-tuned safety set, the ensemble uncertainty is explicitly controlled to/under the specified thresholds.
    \item \textit{Consensus}:
    To effectively utilize the multiple action candidates of the ensemble model while eliminating abrupt changes in action due to the switching system, a consensus system is optimally designed.
    \item \textit{Colored}:
    To facilitate exploration while not interfering with the expert behaviors, autoregressive colored noise is introduced for time-consistent exploration.
\end{enumerate}

\subsection{Controlled ensemble uncertainty}

As mentioned above, EnsembleDAgger has $K$ policy ensemble models $\{\pi_k(a \mid s; \theta_k)\}_{k=1}^K$, each of which is trained with eq.~\eqref{eq:loss_bc}.
Its basic idea is to have diverse inferenced outputs by taking advantage of the fact that $\theta_k$ are initialized differently, and to increase the accuracy of inference statistically.
However, as it is, it is unknown to what extent the differences in output (represented by $\sigma^\pi = \sqrt{K^{-1} \sum_k (a^\pi - a^{\pi_k})^2}$ in EnsembleDAgger), even if the action space is pre-normalized well.
As a result, the threshold $\bar{\sigma}$ for such an ensemble uncertainty is difficult to design in advance and tends to be the wrong $\mathcal{C}$.

\subsubsection{Formulation}

Inspired by the literature~\cite{brantley2020disagreement}, let's consider controlling the ensemble uncertainty during learning.
The proposed modification optimizes the ensemble models not only according to eq.~\eqref{eq:loss_bc}, but also with the following inequality constraint on the ensemble uncertainty.
\begin{align}
    &\ln (2\epsilon) \leq \max_k \ln (\bar{\sigma}^{-1} |a - a^{\pi_k}| + \epsilon) \leq \ln (1 + \epsilon)
\end{align}
where, $\epsilon \ll 1$ denotes the tiny constant that is mainly used for numerical stabilization.
This constraint prevents all $a^{\pi_k}$ from perfectly matching the expert's action $a$, which would impair the statistical performance of the ensemble models, while the upper bound limits the deviation of any action candidate from $a$ to within the threshold.

This inequality constraint is combined with eq.~\eqref{eq:loss_bc} by converting it to the corresponding regularization term~\cite{kobayashi2025lira}.
In summary, the following loss function is minimized.
\begin{align}
    \mathcal{L}^\mathrm{ctrl}(\theta) \!=\! \mathbb{E}_{(s_n, a_n) \sim D} \!\Biggl[\! &
    - \sum_{i=1}^{|\mathcal{A}|} \lambda_i \max_k \ln (\bar{\sigma}^{-1} |a_i - a_i^{\pi_k}| + \epsilon )
    \nonumber \\
    & - \sum_{k=1}^K \ln \pi_k(a_n \mid s_n; \theta_k) \!\Biggr]
\end{align}
where, the subscript $i$ refers to the $i$-th dimension of action, and since the inequality constraint is given on each dimension independently, different Lagrange multipliers, $\lambda_i$, are provided for each.
If the policies are trained well, the inquality constraint should be satisfied correctly.
In other words, it is expected to reduce misjudgments of safety while ensuring the effectiveness of the ensemble models.

\subsubsection{Optimization for satisfying inequality constraint}

$\lambda$ can be optimized as Lagrange multipliers.
\begin{align}
    \phi_\lambda^\ast &= \mathop{\mathrm{arg\,min}}_{\phi_\lambda} \underbrace{- \lambda(s_n; \phi_\lambda) e(a, a^\pi, \delta)}_{\mathcal{L}^\lambda(\phi_\lambda)}
    \\
    e(a, a^\pi, \delta) &= \ln(2\epsilon) + \delta  - \max_k \ln ( \bar{\sigma}^{-1} |a - a^{\pi_k}| + \epsilon )
    \nonumber
\end{align}
where, $\delta \in [0, \ln (1 + \epsilon) - \ln (2\epsilon)]$ denotes the $|\mathcal{A}|$-D slack variable to be optimized later.
Since each state achieves the constraints differently, $\lambda$ is designed as a $|\mathcal{A}|$-D state-dependent function with $\phi_\lambda$ the parameters accordingly.

Finally, the slack variable (the parameters of its state-dependent funciton, $\phi_\delta$) is optimized as follows to assist in satisfying the equality constraint~\cite{kobayashi2025lira}.
\begin{align}
    \delta &= g_\delta(\delta(s_n; \phi_\delta))
    \\
    \mathcal{L}^\delta(\phi_\delta) &=
    \begin{cases}
        \delta(s_n; \phi_\delta) \mathrm{sign}(e(a, a^\pi, \delta)) & |e| > \bar{e}
        \\
        \delta(s_n; \phi_\delta) \lambda & \mathrm{otherwise}
    \end{cases}
\end{align}
where, $g_\delta(\cdot) \in [0, \ln (1 + \epsilon) - \ln (2\epsilon)]$ is the nonlinear transformation to satisfy the domain of the slack variable.
$\bar{e}$ denotes the allowable error, and in this paper, it is designed as 10~\% of the domain size, referring to the literature~\cite{kobayashi2025lira}.
If $e \leq \bar{e}$, the position of equality constraint is shifted towards the lower or upper bound of $\delta$ according to $\lambda$:
if $\lambda > 0$ to maximize the ensemble uncertainty, its behavior is suppressed by making $\delta$ small, and vice versa.
This behavior allows the change in the ensemble uncertainty to be stably controlled while simultaneously searching for solutions that may lie at both boundaries of the inequality constraint.

\subsection{Consensus system among ensemble actions}

Recent DAgger variants, such as EnsembleDAgger, use a safety-based switching system to decide which action $a^c$ to execute from the expert's action candidate $a$ and the agent's action candidate $a^\pi$.
However, as mentioned above, safety decisions left to human are subject to delay, and switching can cause abrupt changes in robot's commands, which can violate the stability of dynamic systems.
Therefore, this study considers the decision of $a^c$ as a kind of consensus-making problem in which a central tendency is derived from multiple candidates, and designs a system that can obtain an optimal consensus with safety taken into account.

\subsubsection{Formulation}

The consensus-making problem of deriving $a^c$ is regarded as the central tendency of multiple action candidates, i.e. $a^{\pi_k}$ ($k=1,\ldots,K$) of the agent and $a$ of the expert.
This can be attributed to the $L_p$ norm minimization problem for the $i$-th action dimension as shown below~\cite{chavent2008central}.
\begin{align}
    a_i^c = \mathop{\mathrm{arg\,min}}_{c_i \in [\underline{a}_i, \overline{a}_i]} \left\{ \sum_{k=1}^K w_k |a_i^{\pi_k} - c_i|^{p_i} + w_{K+1} |a_i - c_i|^{p_i} \right\}^{\frac{1}{p_i}}
\end{align}
where, $\underline{a}_i$ and $\overline{a}_i$ denote the minimum and maximum values among the candidates, respectively.
$w_k \geq 0$ is the weigtht for $k$-th candidate, the sum of which satisfies $1$.
Note that when $a^c$ is inferred only by agents (especially after deployment), the $K+1$-th candidate for $a$ is merely excluded.

With $p_i > 1$, $L_{p_i}$ norm becomes strictly convex, the global optimum is obtained fast and asymptotically.
However, when $p_i$ is very large, Newton methods that require (pseudo-)quadratic gradients become unstable numerically.
Therefore, it is converted to the corresponding root-finding problem for its analytical first-order gradient, which can be asymptotically solved by a kind of bisection methods, so that the numerical stability of the solution is guaranteed.
In practice, ITP method~\cite{oliveira2021enhancement}, which has theoretically shown fast convergence, has been employed in this study.

$a_i^c$ obtained by solving such a problem has different properties according to $p_i$ and $w_k$.
Here, $w_k$ is ignored at once for simplicity, and three representative results are summarized as below.
\begin{itemize}
    \item $p_i \to 1$: This yields the median, which can elimiate outliers mixed in with the candidates, $\mathrm{Quantile(\{a_i^{\pi_k}\}_{k=1}^K \cup \{a\}; 0.5)}$.
    \item $p_i = 2$: This yields the mean, which is the fairest central tendency if the candiates are distributed as Gaussian, $(K+1)^{-1} (\sum_{k=1}^K a_i^{\pi_k} + a)$.
    \item $p_i \to \infty$: This yields the midrange, which is the fairest central tendency if the candidates are distributed uniformly, $(\underline{a}_i + \overline{a}_i)/2$.
\end{itemize}
Note that the intermediate $p_i$ has a mixture of these properties.
As the consensus should be fair while anomalous opinions should be excluded, it is worth desigining the proper $p_i$ according to the distribution shape of candidates.

The above properties are the same even when $w_k$ is included, since it only vary the pseudo-number of candidates.
Thus, given appropriate weights, the conventional discrete switching system can be extended to a continuous one.

\begin{figure}[tb]
    \begin{subfigure}[b]{0.48\linewidth}
        \centering
        \includegraphics[keepaspectratio=true,width=\linewidth]{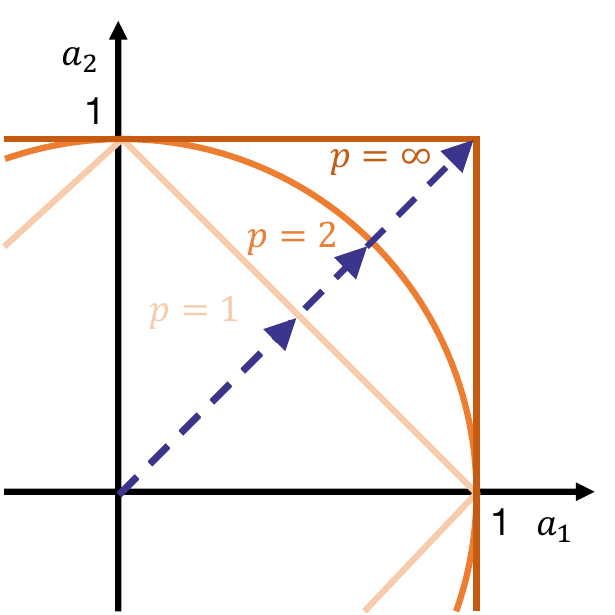}
        \subcaption{Geometry of $L_p$ norm}
        \label{fig:design_p}
    \end{subfigure}
    \begin{subfigure}[b]{0.48\linewidth}
        \centering
        \includegraphics[keepaspectratio=true,width=\linewidth]{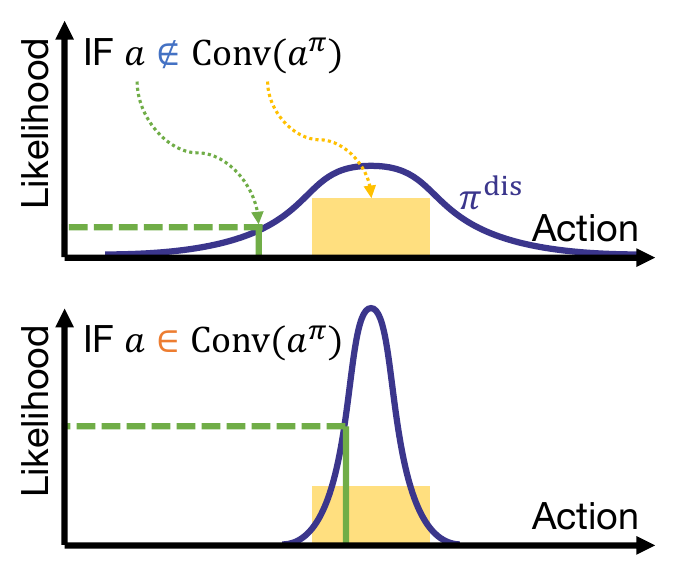}
        \subcaption{Geometry of actions}
        \label{fig:design_w}
    \end{subfigure}
    \caption{Geometries for designing $p_i$ and $w_k$}
    \label{fig:design}
\end{figure}

\subsubsection{Design of $p_i$}

To determine $p_i$, the distribution shape should be quantified, as indicated above.
For this purpose, the ratio of the standard deviation, $\sigma_i^\mathrm{STD}$, and mean absolute error, $\sigma_i^\mathrm{MAE}$, for the candidates is defined as $\rho_i$.
\begin{align}
    \rho_i &= \cfrac{\sigma_i^\mathrm{MAE}}{\sigma_i^\mathrm{STD}}
    \\
    \mathrm{s.t.\ } &\begin{cases}
        \sigma_i^\mathrm{STD} = \sqrt{ \sum_{k=1}^K w_k (\mu_i - a_i^{\pi_k})^2 + w_{K+1} (\mu_i - a_i)^2 }
        \\
        \sigma_i^\mathrm{MAE} = \sum_{k=1}^K w_k |\mu_i - a_i^{\pi_k}| + w_{K+1} |\mu_i - a_i|
        \\
        \mu_i \ \ \ \ = \sum_{k=1}^K w_k a_i^{\pi_k} + w_{K+1} a_i
    \end{cases}
    \nonumber
\end{align}
This ratio is known to be $\sqrt{2/\pi} (\simeq 0.8)$ if the candidates follow a Gaussian distribution.
If the candidates follow a uniform distribution, it asymptotically approaches $1$, and if they are non-Gaussian, such as heavily tailed and/or asymmetric, it approaches $0$.
Therefore, $p_i$ can be defined as the function of $\rho_i$.
However, since $p_i$ and $\rho_i$ have different domains, a reasonable connection between them is required.

This paper focuses on the longest distance $d_i$ from the origin among sets in a $K+1$-dimensional space where the $L_{p_i}$ norm is $1$ (see Fig.~\ref{fig:design_p}).
Three important cases are as follows: $\sqrt{K+1}^{-1}$ for $p_i \to 1$; $1=\sqrt{K+1}^0$ for $p_i = 2$; and $\sqrt{K+1}^1$ for $p_i \to \infty$.
Here, the power part (defined as $q_i$) is bounded by $(-1, 1)$, and the connection with $\rho_i$ can be easily made by the following equation.
\begin{align}
    q_i = 2 \rho_i^{\frac{\ln 2}{\ln \sqrt{\pi/2}}} - 1
\end{align}

Then, the coordinates in the $K+1$-dimensional space with $d_i$ have the same values on all axes, i.e. $d_i / \sqrt{K+1} = \sqrt{K+1}^{q_i-1}$.
With one $L_{p_i}$ norm, $p_i$ is derived as follows:
\begin{align}
    & (K+1) \left( \sqrt{K+1}^{q_i-1} \right)^{p_i} = 1
    \nonumber \\
    \therefore \ & p_i = \frac{1}{1 - \rho_i^{\frac{\ln 2}{\ln \sqrt{\pi/2}}}}
\end{align}

\subsubsection{Design of $w_k$}

The weight $w_k$ ($k=1,\ldots,K+1$) for each candidate is designed.
Each candidate is generated from the corresponding policy $\pi_k$, and its likelihood can be taken as the confidence level.
However, this likelihood is quantifiable only for the agent actions, and cannot be numerically calculated for the expert because $\pi^\ast$ is unknown.
While there may be a direction to receive information equivalent to the likelihood from the expert, or to estimate it, this paper utilizes the safety definition in EnsembleDAgger.

As one of the safety with the threshold $\bar{\sigma}$ for the ensemble uncertainty, an alternative policy $\pi^\mathrm{exp}$ to $\pi^\ast$ is assumed with a diagonal normal distribution with $a$ the mean and $\bar{\sigma}/3$ the scale parameters.
This means that the likelihood of $\pi_k$ is essentially smaller than that of $\pi^\mathrm{exp}$ unless the ensemble uncertainty is controlled to be below the threshold and the scale of $\pi_k$ is correspondingly smaller, resulting in a preference for the expert's action $a$.
Another safety is that the difference between $a$ and $\mu^\pi = K^{-1}\sum_{k=1}^K a^{\pi_k}$ should be below a threshold.
This can be represented with another diagonal normal distribution with $\mu^\pi$ the mean and the mean squared error of $a$ and $a^{\pi_k}$ as its variance, $\pi^\mathrm{dis}$ (see Fig.~\ref{fig:design_w}).
That is, if the likelihood of $a$ of $\pi^\mathrm{dis}$ is large, the agent would be similar to the expert and safe.
Note that $\bar{\Delta}$ in EnsembleDAgger is not needed since $\pi^\mathrm{dis}$ makes it easier to confirm whether $a$ is contained inside the convex hull given by $a^{\pi_k}$, leading to $a^c \simeq a$ even without $a$.

Finally, the weight $w_k$ is designed as follows:
\begin{align}
    w_k \propto
    \begin{cases}
        \cfrac{\pi^\mathrm{exp}(a \mid a, \bar{\sigma}/3)}{\pi^\mathrm{dis}(a \mid \mu^\pi, \sqrt{K^{-1} \sum_{k=1}^K (a - a^{\pi_k})^2})} & k = K+1
        \\
        \pi_k(a^{\pi_k} \mid s; \theta_k) & \mathrm{otherwise}
    \end{cases}
\end{align}
Note that $w_k$ is normalized to sum to $1$ to satisfy its definition.
Thanks to this design, it allows for safety considerations analogous to EnsembleDAgger, while obtaining continuous $a^c$ decisions than simple switching systems.

\subsection{Colored noise for time-consistent exploration}

In the context of RL, action noise from the stochastic policy is effective, and DART variants~\cite{laskey2017dart,oh2023bayesian}, another approach to IIL, make the trained policy robust by adding optimized noise to the expert's behavior.
Thus, the exploration noise is also considered effective even in DAgger variants.
In the above research cases (with continuous action space), however, the exploration noise is white noise generated from a Gaussian distribution.
In other words, noise is time-independent, and a robot system fastly oscillated by it might be difficult for a human expert to operate (see the attached video and the section~\ref{subsec:exp}).

\begin{figure}[tb]
    \centering
    \includegraphics[keepaspectratio=true,width=0.96\linewidth]{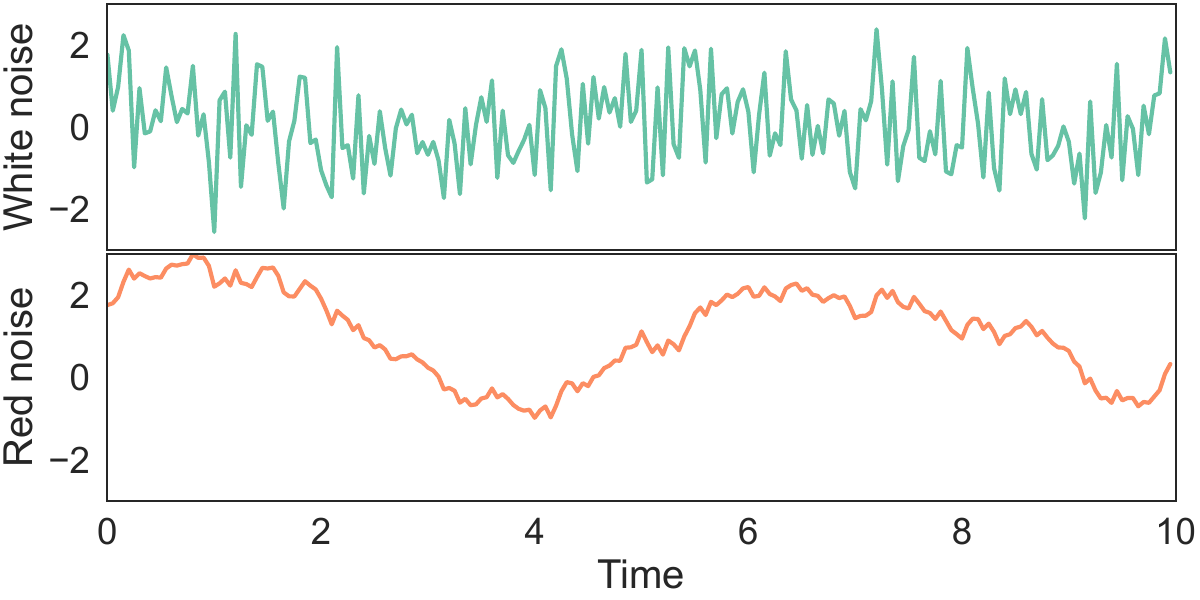}
    \caption{Time consistency of red noise ($\Delta t=0.05$ and $T=3$)}
    \label{fig:noise}
\end{figure}

Therefore, this study employes colored noise, which has been reported to accelerate exploration rather than white one, inspired by the literature~\cite{eberhard2023pink}.
However, previous studies have used an implementation that generates time-series noises with a specified time step in advance, perhaps in order to investigate general colored noise, which is not flexible enough in practice.
Red noise is therefore selected since it can be generated with an autoregressive model as follows:
\begin{align}
    \epsilon^\mathrm{red}_t = \gamma \epsilon^\mathrm{red}_{t-1} + \sqrt{1 - \gamma^2} \epsilon^\mathrm{white}_t
\end{align}
where $\epsilon^\mathrm{red}_{0} = \epsilon^\mathrm{white}_{0}$.
$\gamma \in [0, 1)$ represents the temporal consistency, and is given as $\exp(-\Delta t / T)$ with $\Delta t$ the time step period and $T$ the time constant (see Fig.~\ref{fig:noise}).

To add the red noise into the agent's actions, the likelihood (a.k.a. confidence for computing $w_k$) needs to be taken into account.
That is, the excessive red noise to prioritize exploration would make the likelihood smaller than others (especially the expert's one), losing the chance of reflecting it to $a^c$.
With this in mind, $a^\pi$ is finally given as follows:
\begin{align}
    a^{\pi_k} = \mu_{\pi_k} + \cfrac{2\sqrt{K}}{3} \sigma_{\pi_k} \epsilon^\mathrm{red}_{t,k}
\end{align}
where, $\mu_{\pi_k}$ and $\sigma_{\pi_k}$ denote the mean and scale of $\pi_k$, which is modeled as a diagonal normal distribution, respectively.
Note that the component-wise $\epsilon^\mathrm{red}_{t,k}$ is different for each dimension of the action space.
As white and red noises are generally within $\pm 2 \sim 3$, this gain, $2\sqrt{K}/3$, increases the expert's confidence level relatively only to $K \sim 2K$, even if all components carry the maximum amount of exploration noise.
In other words, the difference of the number of candidates from the agent and expert is compensated with this design, and the actions from the agent are possibly reflected into $a^c$ (depending on $p_i$ and $w_k$).

\section{Experiments}

\begin{figure*}[tb]
    \centering
    \includegraphics[keepaspectratio=true,width=0.96\linewidth]{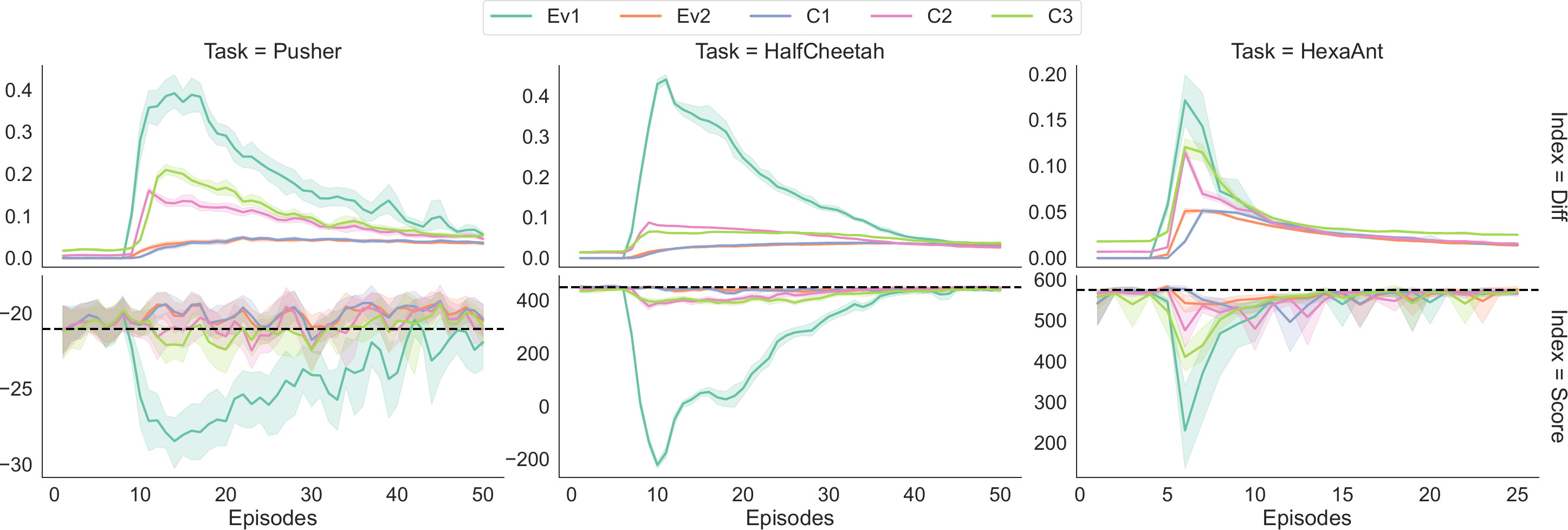}
    \caption{Trajectories during data collection (dashed lines: the experts' average scores)}
    \label{fig:sim_learn}
\end{figure*}

\subsection{Simulations}

In order to statistically verify the performance of the proposed method, CubeDAgger, the following simulations are conducted.
The important criteria are threefold:
i) the difference between the expert's action $a$ and the executed action $a^c$ during the data collection process;
ii) the degree of degradation of control performance during the data collection process;
iii) the robustness of the trained policies.

\subsubsection{Tasks}

Three dynamic tasks with different robots simulated in Mujoco are performed.
The first \textit{Pusher}, where a manipulator pushes an object to its destination without grasping it, might cause the object to slide into hard-to-push positions by wrong actions.
The second \textit{HalfCheetah}, where a 2D legged robot moves forward while standing on its head%
\footnote{This behavior was achieved by changing the weights of rewards},
is unstable and sensitive to mistakes.
The third \textit{HexaAnt}, which has a multi-degree-of-freedom system with three joints in each of its six legs for walking foward, is easy to intervene in state transitions and stability with various action patterns.
These tasks were trained multiple times with the RL algorithm~\cite{kobayashi2024revisiting}, and the best policies are selected as their experts.
Since the experts maximized the sum of rewards, the control performance is evaluated by it.

\subsubsection{Models}

The policy of the imitation agent is approximated by neural networks with two fully-connected layers (each with 100 neurons) in the hidden layer.
To make it a lightweight ensemble model, only the output head is separated to take $K=10$ outputs (i.e. the mean and log scale of Gaussian distribution), while sharing the input and hidden layers~\cite{osband2016deep}.
This policy ensembles (and other learnable parameters) are optimized using AdaTerm~\cite{ilboudo2023adaterm}, a stochastic gradient descent method that is robust to noise and outliers, with its default configuration.
The dataset $D$ has the sufficient capacity to hold all the data during the experiments, and all the data in $D$ are uniformly randomly replayed once at the end of each episode with batch size $B=50$ each.

\subsubsection{Comparisons}

The following five conditions including ablation tests are compared.
\begin{itemize}
    \item \textit{EV1}: EnsembleDAgger with (almost) recommended configurations, i.e. $\bar{\Delta} = 1$ and $\bar{\sigma} = 0.1$.
    \item \textit{EV2}: EnsembleDAgger that eliminates $\bar{\Delta}$ as in the proposed method, $\bar{\Delta} = 0.1$ and $\bar{\sigma} = 0.1$.
    \item \textit{C1}: \textit{EV2} with the controlled ensemble uncertainty.
    \item \textit{C2}: \textit{C1} with the consensus system.
    \item \textit{C3}: The proposed method with all the modifications including the colored noise (a.k.a. CubeDAgger).
\end{itemize}

\subsubsection{Results}

The difference $|a - a^c|$ (labeled \textit{Diff}) and the sum of rewards (labeled \textit{Score}) at the time of data collection are plotted in Fig.~\ref{fig:sim_learn}.
Apparently, \textit{EV1} frequently performed actions different from those of the experts, causing large drops in the scores.
This behavior would involve significant risk during data collection in real-world robot applications, even if it may be needed to learn how to recover from (accumulated) errors in case of failure.
On the other hand, \textit{EV2} mostly prioritized $a$ for determining $a^c$, and the scores were always high.
In other words, this behavior suggests a lack of exploration, although this will be evident later when we look at the behavior of the trained policies.

Focusing on the proposed modifications, \textit{C1} did not change much from \textit{EV2}, except for a slight delay in the timing at which $|a - a^c|$ began to grow.
On the other hand, the behavior of \textit{C2} was clearly different, and its exploratory behavior was between those of \textit{EV1} and \textit{EV2}.
The fact that the score did not decrease much suggests that the consensus was gained more safely than the case with EnsembleDAgger (i.e. the simple if-then rule).
In addition, although $|a - a^c|$ should become smaller as the episodes pass due to the convergence of the agent policy to the expert's one, \textit{C3} made it remain for longer period thanks to the colored noise moderately added.
The noise also caused a slight decrease in the score, but not as much as in the case of \textit{EV1}.

To compare the five methods more quantitatively, two criteria are statistically evaluated.
The one shows the ability to maintain the control performance during data collection, labeled \textit{Retention}, which is computed as the mean of scores during data collection.
The other shows the robustness of the trained policies, labeled \textit{Robustness}, which is evaluated by deploying them under the disturbed environments, where a uniform random disturbance with a probability of 5~\% to each action dimension.
Note that the maximum intensity of the disturbance was set to the largest action of each dimension for \textit{Pusher} and \textit{HexaAnt}, and to the half of it for \textit{HalfCheetah} due to its instability.

\begin{figure}[tb]
    \centering
    \includegraphics[keepaspectratio=true,width=0.96\linewidth]{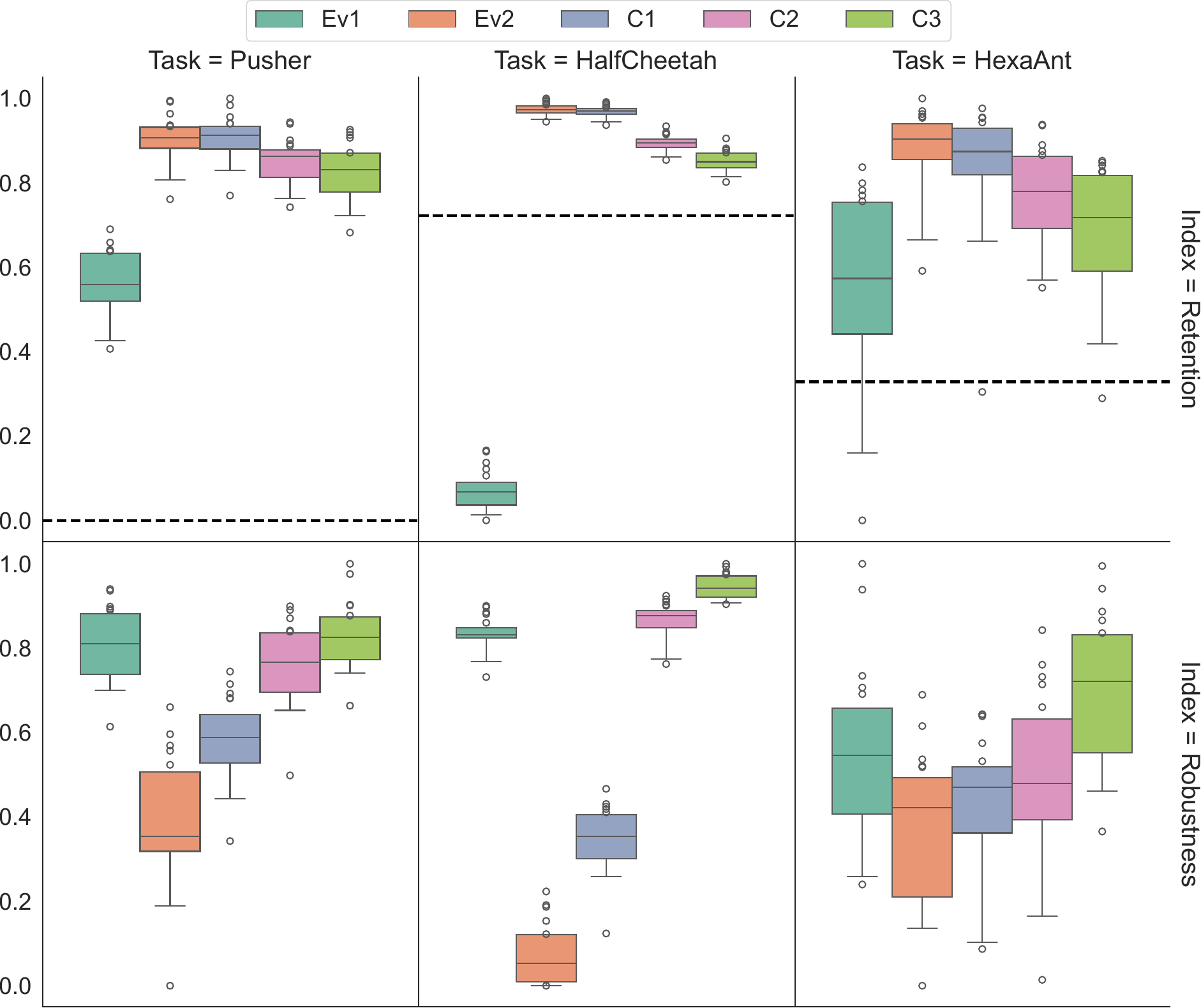}
    \caption{Statistical evaluation with 21 random seeds for each condition (dashed lines: the experts' worst-case scores)}
    \label{fig:sim_eval}
\end{figure}

The normalized results of both criteria are summarized in Fig.~\ref{fig:sim_eval}.
As the learning process suggested, \textit{EV1} caused a significant performance degradation, while \textit{EV2} maintained the expert's control performance.
Although the control performance degraded with each proposed modification, it rarely exceeded the worst case of experts (given as the dashed lines).
It would be expected that the robustness should be inversely proportional to the retention, and this is the case when looking at \textit{EV1} and \textit{EV2}.
However, with the similar retention to \textit{EV2}, \textit{C1} was clearly more robust.
More surprisingly, \textit{C2}, which has the better retention than \textit{EV1}, achieved the same level of robustness, and \textit{C3} achieved even better robustness.
This indicates that the proposed CubeDAgger could more efficiently facilitate the exploration to improve the robustness without violating the dynamic stability.

\subsection{Scooping balls with quadruped robot}
\label{subsec:exp}

To demonstrate the practicality of CubeDAgger, real-world experiments are conducted.
Since a human serves as the expert, the following three points are confirmed:
i) EnsembleDAgger's switching system hinders proper human supervision;
ii) Exploration with time-inconsistent white noise impairs human operability;
iii) CubeDAgger mitigates these issues and enables stable imitation from human.

\subsubsection{Conditions}

The experiments use a quadruped robot developed by Ahead.IO (see Fig.~\ref{fig:scoop_snap}).
Its torso pose is kinematically controlled, so the action space is defined as six dimensions for its target.
Note that the action is sent to the robot 30~Hz and smoothed in the internal controller, which runs 90~Hz.
A net is attached to the lower of the torso, with upper and lower boundaries of the action space established to ensure the net reached the ground and the entire workspace.
A Intel RealSense D435i is mounted at the front of the torso to capture 640x480 RGB images at 30~Hz.
The human expert views them and operate the robot via a SpaceMouse.

At the start of each episode, three balls are positioned around the net within the range captured.
They are scooped in the order of front, right, and left, although this order is not necessarily satisfied due to disturbances caused by the agent's actions.
Each episode is set to one minute (i.e. 1800 steps).
Learning is completed over 30 episodes (i.e. 30 minutes), with the agent's policy being evaluated every five episodes.

To accomplish this task, its state space is defined as follows.
First, the six-dimensional target pose in the internal controller is required to predict the next target pose according to the given action.
Second, the RGB image is reshaped to 518x518 and converted to the corresponding 384-dimensional visual features using DINOv2~\cite{oquabdinov2} (more specifically, \texttt{vit\_small\_patch14\_dinov2.lvd142m} in Hugging Face).
Finally, to perceive the dynamical features, the above two are fed into an echo state network~\cite{jaeger2002adaptive} with 512 neurons.
In total, the state space has 902 dimensions.

The model and learning conditions for the agent's policy are mostly the same as those in the simulations, but the following three points differ.
To avoid underfitting with the larger state dimension, the neurons of the two hidden layers are increased to 512 and 256, respectively.
To carefully adapt to more diverse situations, the learning rate of AdaTerm is reduced $1/10$ times (i.e. $10^{-4}$).
To achieve stable real-time control at 30~Hz, the number of optimization iterations of ITP method is reduced by half (i.e. eight).

\subsubsection{Results}

\begin{figure}[tb]
    \centering
    \includegraphics[keepaspectratio=true,width=0.96\linewidth]{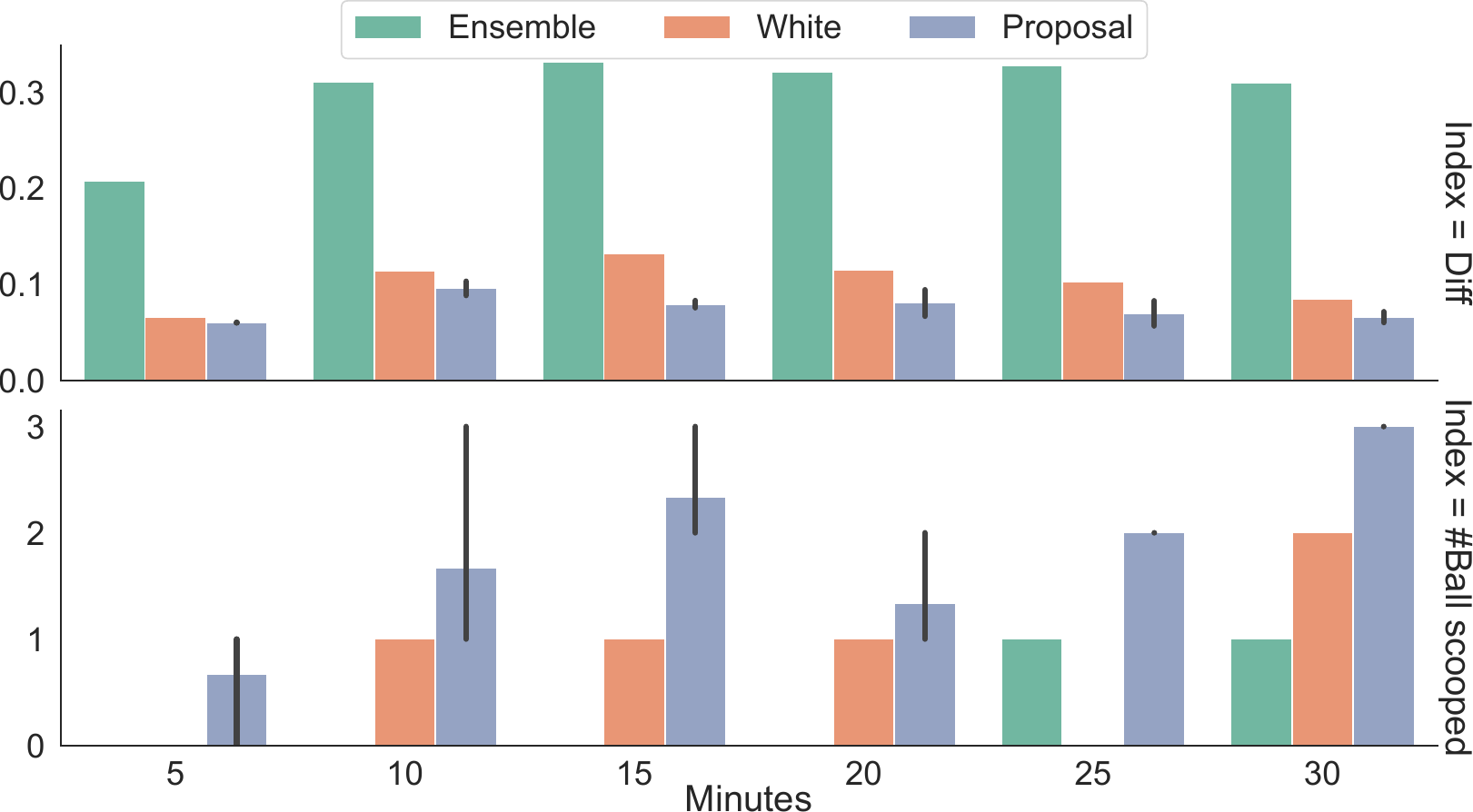}
    \caption{Evaluation of real-world scooping task}
    \label{fig:exp_eval}
\end{figure}

The experiments are summarized in the attached video.
First, with EnsembleDAgger with $\bar{\Delta} = 1$ and $\bar{\sigma} = 0.1$, the agent monopolized control authority early in episodes, making erroneous motions.
To gain the control authority, the human expert intentionally gave excessive actions, but the actions suddenly switched caused such as collisions with walls due to the delay in handling them.

CubeDAgger with white noise for exploration (i.e. $T\to0$) could eliminate the dangerous switching behaviors.
However, the oscillatory behaviors exhibited by the agent made it difficult for the human expert to provide optimal operation.
Furthermore, such oscillations affected the observed images; the screen, constantly fluctuating minutely, caused motion sickness in the human expert.

In contrast, the proposed CubeDAgger successfully executed all episodes safely while promoting diversity in situations through moderate perturbations.
While the above two baselines were truncated only with one trial for each due to heavy burden on the human expert, the proposed method could be conducted three trials comfortably.
Indeed, only two balls could not be scooped in total episodes.

More quantitatively, the test results of the agent policies conducted every five episodes are depicted in Fig.~\ref{fig:exp_eval}.
Consistent with subjective ease of operation, the average $|a - a^c|$ every 5 minutes was smallest with the proposed method.
In addition, the baselines never scoop all three balls even once, whereas the proposed method eventually succeeded in scooping all balls in all three trials.

Furthermore, when the acquired policies by the proposed method were tested five times, the success rate of scooping the three balls was $8/15$.
Remarkably, Among the successful eight episodes, some performed error recovery motions, indicating the benefits of IIL.
In the failed seven episodes, while two balls could be scooped, most cases involved misrecogniting the state as allowing three balls to be scooped: namely, the last ball was occluded behind the net in the images.
Alternatively, the unusual orders mixed in with the supervision data were executed, leading to the situations where the policies were insufficiently optimized.

\section{Conclusion}

This paper proposed a novel IIL method, named CubeDAgger, an improved version of EnsembleDAgger, for making it applicable to dynamic tasks.
The improvements are threefold:
i) control the output variance of the ensemble model to make the safety decision work better;
ii) optimize a consensus problem for multiple action candidates; and
iii) add colored noise for efficient stochastic exploration.
With these improvements, CubeDAgger successfully improved the robustness of trained policies while maintaining dynamic stability during data collection, compared to EnsembleDAgger.
In the real-world scooping demonstration, CubeDAgger successfully acquired robust policies while safely collecting data without excessively interfering with the human operator.

However, as found in the experiments, human experts in dynamic systems cannot always select optimal actions, leading to noisy data.
Noise-robust learning algorithms and/or data selection methods might mitigate this adverse effect.
Then, the limitations of CubeDAgger will be investigated by scaling up to multi-task and more dynamic problems.
Meanwhile, the proposed consensus mechanism could be applicable to shared control, which is worth exploring.

\bibliographystyle{IEEEtran}
{
\bibliography{biblio}
}


\end{document}